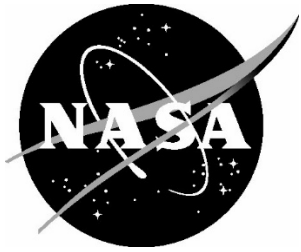

# *Collaborative Agent Reasoning Engineering* (CARE): A Structured Three-Party Design Methodology for Systematically Engineering AI Agents with SMEs, Developers, and Helper Agents

*Rahul Ramachandran*
*Marshall Space Flight Center, Huntsville, Alabama*

*Nidhi Jha*
*The University of Alabama in Huntsville, Huntsville, Alabama*

*Muthukumaran Ramasubramanian*
*The University of Alabama in Huntsville, Huntsville, Alabama*

# NASA STI Program Report Series

Since its founding, NASA has been dedicated to the advancement of aeronautics and space science. The NASA scientific and technical information (STI) program plays a key part in helping NASA maintain this important role.

The NASA STI program operates under the auspices of the Agency Chief Information Officer. It collects, organizes, provides for archiving, and disseminates NASA's STI. The NASA STI program provides access to the NTRS Registered and its public interface, the NASA Technical Reports Server, thus providing one of the largest collections of aeronautical and space science STI in the world. Results are published in both non-NASA channels and by NASA in the NASA STI Report Series, which includes the following report types:

- TECHNICAL PUBLICATION. Reports of completed research or a major significant phase of research that present the results of NASA Programs and include extensive data or theoretical analysis. Includes compilations of significant scientific and technical data and information deemed to be of continuing reference value. NASA counterpart of peer-reviewed formal professional papers but has less stringent limitations on manuscript length and extent of graphic presentations.

- TECHNICAL MEMORANDUM. Scientific and technical findings that are preliminary or of specialized interest, e.g., quick release reports, working papers, and bibliographies that contain minimal annotation. Does not contain extensive analysis.

- CONTRACTOR REPORT. Scientific and technical findings by NASA-sponsored contractors and grantees.

- CONFERENCE PUBLICATION. Collected papers from scientific and technical conferences, symposia, seminars, or other meetings sponsored or co-sponsored by NASA.

- SPECIAL PUBLICATION. Scientific, technical, or historical information from NASA programs, projects, and missions, often concerned with subjects having substantial public interest.

- TECHNICAL TRANSLATION. English-language translations of foreign scientific and technical material pertinent to NASA's mission.

Specialized services also include organizing and publishing research results, distributing specialized research announcements and feeds, providing information desk and personal search support, and enabling data exchange services.

For more information about the NASA STI program, see the following:

- Access the NASA STI program home page at http://www.sti.nasa.gov

.

NASA/TM–20260000926

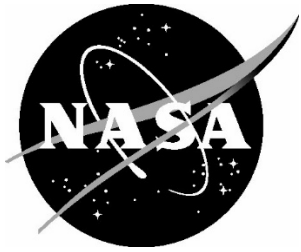

# *Collaborative Agent Reasoning Engineering* (CARE): A Structured Three-Party Design Methodology for Systematically Engineering AI Agents with SMEs, Developers, and Helper Agents

*Rahul Ramachandran*
*Marshall Space Flight Center, Huntsville, Alabama*

*Nidhi Jha*
*The University of Alabama in Huntsville, Huntsville, Alabama*

*Muthukumaran Ramasubramanian*
*The University of Alabama in Huntsville, Huntsville, Alabama*

# Abstract

This technology memo describes CARE (Collaborative Agent Reasoning Engineering), a disciplined, stage-gated process for engineering large language model (LLM) agents that specifies behavior, grounding, tool orchestration, and verification through reusable artifacts rather than trial-and-error prompt iteration. CARE reframes agent building as an engineering discipline centered on explicit specifications. The process is defined as a three-party workflow among subject-matter experts (SMEs), developers, and required LLM-based helper agents, where helper agents facilitate conversion of informal domain intent into structured, reviewable artifacts that humans approve at stage gates.

CARE is motivated by the uneven performance landscape of LLMs, where results vary significantly based on a user's ability to manage domain-specific constraints and verification steps, often widening the gap between novice and expert analysts. CARE targets this practical reliability gap that emerges from uneven LLM performance and user application. CARE operationalizes agent development by producing concrete artifacts: interaction requirements, domain-grounding specifications, tool interfaces and orchestration plans, reasoning policies and guardrails, prompt architecture, and evaluation criteria, so that agent behavior is specifiable, testable, and maintainable over time.

In a case study building a NASA Earth science data discovery agent that queries the NASA Common Metadata Repository (CMR) Application Program Interface (API), the CARE-designed agent outperformed a baseline with identical model and tool access in a two-gate evaluation achieving higher Recall@1 on a large synthetic benchmark (n=621; 71.7% vs 69.1%) and higher Recall@5 on an SME-created gold benchmark (n=43; 27.2% vs 20.2%). CARE shows measurable retrieval gains under both fast synthetic testing and SME-grounded verification. Thus, CARE is presented as a stage-gated, artifact-driven, three-party methodology involving helper agents that yields measurable performance improvements in a realistic, tool-driven scientific data discovery setting. CARE demonstrates a repeatable method with concrete evidence of improved outcomes.

## 1. Introduction

Large language models (LLMs) now produce high-quality text and code across many tasks, which makes them attractive for scientific and technical knowledge work where analysts must interpret requirements, retrieve information, and synthesize evidence into decisions. LLMs create a credible opportunity to accelerate scientific and technical workflows. Yet LLM performance remains inconsistent across tasks, a pattern often described as a "jagged technological frontier," where outputs can be highly useful within certain task regimes but unreliable outside them, and where user skill strongly influences whether productivity improves or degrades. The central challenge is not only capability, but also dependable use within shifting boundaries. This unevenness produces a practical gap between novice and expert analysts, because experts more consistently apply domain constraints, verify intermediate outputs, and structure multi-step workflows, while novices are more likely to accept fluent but incorrect results or to under-specify intent and constraints.

Custom LLM agents offer a path to reduce this gap by embedding expert analytic practices into reusable system behavior, including disciplined decomposition, verification, domain grounding, and tool-using workflows that make retrieval and computation authoritative rather than speculative. Agents can operationalize expert workflows so performance is less dependent on individual user skill. However, building reliable agents is non-trivial, because behavior emerges from the interaction among prompts, grounding content, tools, orchestration logic, and evaluation criteria. Additionally, trial-and-error design can yield systems that fail silently, regress after small changes, or overfit to narrow test cases.

Robust agents require an engineering discipline that makes intent explicit and verifiable. Traditional software processes support iteration, but they do not directly solve agent-engineering problems such as specifying intended reasoning behavior, translating SME knowledge into stable grounding artifacts, designing guardrails for uncertainty and failure modes, and constructing verification benchmarks that detect regressions over time. What's missing is a methodology tailored to specifying and verifying LLM-agent behavior.

This paper proposes CARE (Collaborative Agent Reasoning Engineering), a disciplined, stage-gated process for engineering LLM agents in which subject-matter experts, developers, and required LLM-based helper agents work together to specify behavior and processes through reusable artifacts rather than ad hoc prompt iteration. In the remainder of the paper, we refer to these LLM-based helper agents simply as "helper agents." The idea of helper agents is influenced by the accelerated knowledge discovery (AKD) concept [7], which proposes AI as a tool to augment the cognitive capabilities of scientists to help make knowledge discoveries faster. We extend this idea beyond scientific tasks to the agent design process itself by using helper agents to augment the specification and engineering workflow. Thus, the helper agents are framed as an application of AKD-style cognitive augmentation to the engineering of agents.

In CARE, helper agents function as facilitation infrastructure, converting informal discussion into structured, reviewable artifacts by asking targeted, phase-aligned questions, drafting and normalizing specifications, and proposing concrete revisions that humans approve or reject at defined stage gates. Helper agents provide "specification throughput" while humans retain procedural control.

CARE produces concrete deliverables including interaction requirements, domain-grounding specifications, tool interfaces and orchestration plans, reasoning policies and guardrails, prompt architecture, and evaluation criteria. With these deliverables, agent behavior can be reviewed, tested, iterated, and maintained as models, tools, and requirements evolve.

In this paper, we make three contributions: (1) we introduce CARE as a stage-gated, artifact-driven methodology for engineering LLM agents via triadic collaboration among SMEs, developers, and helper agents; (2) we deconstruct agentic functions into interaction policy, domain grounding, tool orchestration, and evaluation/verification and map these targets to CARE artifacts and gates; and (3) we present a NASA Earth science data discovery case study using the NASA CMR API, evaluating via a two-gate benchmarking approach with Recall@K metrics, showing that a CARE-designed agent outperforms a baseline with identical model and tool access.

## 2. Background

Knowledge-intensive scientific and technical work is typically organized as workflows rather than isolated tasks, where analysts must translate objectives into sub-questions, retrieve and validate external evidence, apply domain constraints, and communicate results in forms that others can validate and reuse. LLMs can accelerate parts of these workflows, but prior work characterizes their capabilities as uneven across tasks and contexts, which makes outcomes hard to predict and dependent on a user's skill level with LLM use [1]. In practice, expert analysts often succeed not only because they ask better questions, but also because they consistently apply domain constraints, select appropriate tools, verify intermediate outputs, and recognize when a task is outside a model's reliable regime, while novice analysts are more likely to under-specify intent or accept convincing but incorrect results.

This novice-expert gap motivates LLM-based agents as a systems approach, where the goal is to encode an expert workflow by combining (i) an interaction policy that guides decomposition and verification, (ii) domain grounding that supplies stable reference context, and (iii) tool use that provides authoritative retrieval and computation. Agents serve as a mechanism for turning tacit expert workflows into reusable system behavior.

Operationalizing expert workflows at scale requires a repeatable way to externalize experiential knowledge into specifications, and CARE assumes this is best achieved through structured collaboration in which helper agents support the translation of informal intent into explicit artifacts under human review. Agent behavior is an emergent property of multiple interacting elements such as prompts, grounding documents, tool APIs, orchestration logic, and evaluation harnesses, so small changes in context, model versions, or tools can cause silent regressions that are difficult to detect without custom testing. These issues make agentic systems difficult to maintain without explicit artifacts and verification.

Scientific and technical deployments also face distinct risk modes, including hallucinated claims presented with undue confidence, tool misuse or mis-parameterization, incorrect aggregation of retrieved evidence, and "silent failures" where outputs look plausible but violate domain constraints or provenance expectations. Reliability requires designing for controlled failure behavior, not just improving average-case output quality. These constraints arise across common task classes such as literature review and synthesis, gap analysis, dataset search, code search, and full analysis workflows, where correctness, traceability, and reproducibility are paramount.

Consequently, the central problem addressed in this paper is methodological. Teams need a collaborative way for SMEs and developers to design agentic processes and elements that can be reviewed and tested, and to specify evaluation criteria that verify performance on realistic complex queries rather than curated demos. CARE methodology is proposed as that process, explicitly combining requirements capture for constraints and objectives with reasoning/prompt and tool-orchestration engineering, and attaching stage gates that make progress measurable and regressions detectable as the agent evolves.

## 3. Related Work

Prior work applying LLMs in knowledge work emphasizes that capability gains are real but uneven, and that outcomes depend strongly on whether users can structure tasks, verify outputs, and remain within regimes where model behavior is dependable, motivating approaches that make expert-like workflows more accessible to novices [1]. A growing body of work on agentic LLM systems demonstrates that multi-step workflows can be executed by combining task decomposition, iterative refinement, and (when applicable) tool use, strengthening the feasibility case for using agents to operationalize complex processes rather than single-shot responses. [4,5,6]. These systems establish feasibility of agentic workflows but do not, by themselves, specify how teams should engineer them systematically.

Separately, engineering approaches for tool-using assistants provide practical scaffolding for retrieval, API calling, and orchestration. However, these efforts typically emphasize runtime infrastructure and integration patterns rather than collaborative pre-implementation specification of behavior, grounding, and verification that remains stable across iterations and model/tool drift. These infrastructure solutions do not solve issues associated with specifying and correctly verifying the agent.

Furthermore, prompting and interaction-pattern work [2] provides reusable structures for decomposition, tool use, self-checking, and verification behaviors, yet these patterns are commonly applied through iterative tinkering without explicit requirements, shared artifacts, or stage gates that allow SMEs and developers to review intent and detect regressions. These prompt patterns [2] are useful, but without artifacts and gates they tend to remain ad hoc and hard to maintain.

Requirements engineering and participatory design have long studied how to elicit expert knowledge, translate stakeholder goals into implementable specifications, and validate systems against acceptance criteria, but design of LLM based agents introduce their own distinct challenges because behavior can shift with model versions, prompt/context updates, and tool availability. These potential pitfalls increase the need for artifact versioning, explicit guardrails, and verification designs that detect silent failures. CARE builds on established specification disciplines while adapting them to the brittleness and emergent behavior of LLM agents. Evaluating LLM agents has shown that benchmark choice and scoring rubrics strongly affect conclusions, and that "demo success" can mask failure on realistic, ambiguous, or high-constraint tasks, supporting the idea that verification must be designed alongside the agent rather than retrofitted after implementation.

Separately, accelerated knowledge discovery (AKD) frames AI as a tool to augment the cognitive capabilities of scientists and accelerate discovery, and CARE extends this augmentation idea upstream to the engineering process by using helper agents to accelerate elicitation and artifact production under human review [7]. Thus, CARE is a methodology that treats helper agents as partners in converting informal SME intent into structured, reviewable specifications and candidate evaluation artifacts under human approval. CARE is not a new runtime agent framework, but a three-party, artifact-centered engineering creation process that makes design intent verifiable and maintainable.

# 4. Deconstructing an LLM Agent

An LLM "agent" is best understood as a system that repeatedly transforms an input goal into intermediate decisions and actions, rather than as a single prompt that produces a single response. This perspective means that agent quality depends on how the system structures reasoning, uses information, executes tools, and validates outcomes across multiple steps. As such, agents should be treated as multi-step systems, not single-shot prompt responders, and these systems should be constructed with the following four design targets in mind.

## 4.1 Interaction Policy and Reasoning Strategy

The first design target is interaction policy and reasoning strategy, which specifies how the agent interprets intent, decomposes tasks, sequences reasoning operations, manages uncertainty, and decides when to ask clarifying questions versus when to proceed with tool use or synthesis. The behavior design is captured in this reasoning-policy specification. The interaction policy is where expert practices are embedded, because it encodes process discipline that experts apply implicitly such as verification habits, constraint checking, and structured decomposition. These practices then become reusable system behavior rather than manifestations of individual skill. The novice–expert gap is addressed here by making the expert process explicit in the agent's policy.

## 4.2 Domain Grounding

Domain grounding determines what knowledge the agent should treat as authoritative and how it should use that knowledge to constrain interpretation and reduce plausible-but-wrong outputs. The aim is to adhere to valid terminology, assumptions, schemas, and decision criteria within an identified domain. Grounding defines the authoritative domain boundaries that stabilize the agent's behavior.

## 4.3 Tool Orchestration

The third design target is tool orchestration, which specifies what external tools the agent may use, what each tool is for, how inputs and outputs are represented, how the agent chooses among tools, and how it handles ambiguity, retries, errors, and partial results, whether via linear chains or conditional branching. Tool orchestration converts reasoning into robust, executable workflows. Consequently, tool orchestration improves reliability only when the agent invokes tools at appropriate times, validates tool outputs, and carries provenance forward into final outputs, rather than substituting ungrounded generation for authoritative retrieval or computation.

## 4.4 Evaluation and Verification

The fourth design target is evaluation and verification, This target defines what "success" means for the agent from the perspective of the users, how performance is measured on realistic tasks (including complex and ambiguous queries), and how regressions are detected as prompts, context, models, or tools change over time. Evaluating the system against relevant challenges ensures the agent remains accurate even as the technology behind it evolves.

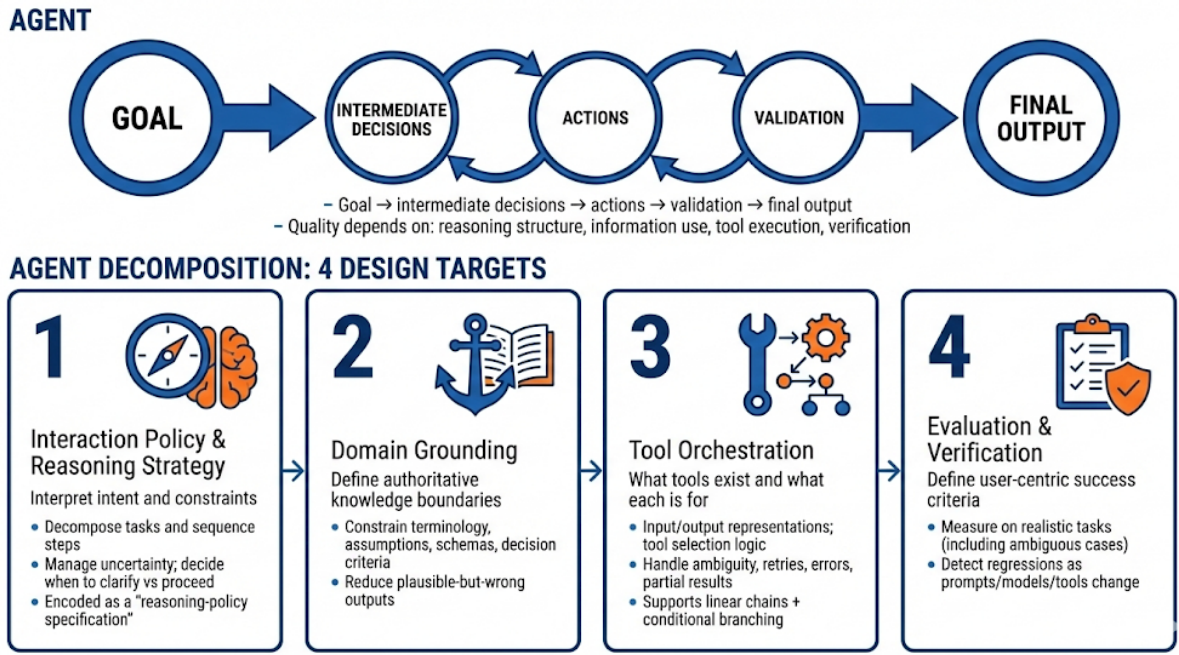

**Figure 1:** Agent decomposition

These targets interact, meaning failures often arise at their boundaries. For example, failures can occur when correct tools are used under an incorrect reasoning policy, correct grounding is ignored during synthesis, or apparent success disappears once verification moves from demos to realistic benchmarks. Disciplined engineering requires artifacts that specify each target and their interfaces. Such systematic decomposition clarifies why trial-and-error prompt tweaking cannot reliably produce robust agents.

CARE builds on this deconstruction by mapping each design target to concrete artifacts and stage gates, enabling systematic refinement while keeping progress measurable and regressions detectable as the agent and its environment evolve. The design targets provide the conceptual scaffold that CARE operationalizes.

## 5. CARE Methodology

### 5.1 Overview

In the Collaborative Agent Reasoning Engineering (CARE) collaborative methodology for engineering LLM agents, subject-matter experts (SMEs), developers, and LLM-based helper agents jointly translate domain expertise and workflow requirements into explicit, reviewable artifacts, rather than relying on iterative prompt tinkering that is difficult to reproduce and verify. CARE reframes agent building as collaborative engineering with concrete, reviewable deliverables. CARE is organized into systematic phases with stage gates, where each phase produces artifacts that specify the design targets described previously, so that agent behavior can be tested and maintained even as assets and prompts change. The overall process is staged, starting from broad scope and requirements, then gathering details (context, tools, output format, and guardrails) before designing the reasoning document. Only after the reasoning document passes the review gate does the team design the agent prompt.

**Phase 1: Scope and Decompose the Agent** This phase establishes the target workflow, intended users, and constraints, and uses the agent deconstruction to define what must be specified for interaction policy, grounding, orchestration, and verification before implementation begins. In this phase, the helper agent drafts an initial scoped artifact set (constraints, decomposition targets, intended users) to seed SME and developer review, converting an "agent idea" into explicit design targets and constraints. Phase 1 thus produces a shared, reviewable specification starting point.

**Phase 2: Key Information Elicitation (Steps 2.1–2.3)** In this phase, key inputs needed to encode expert practice are captured by focusing on (2.1) tools, (2.2) context, and (2.3) output format, so that the agent's operational environment and expected deliverables are explicitly specified before proceeding into detailed behavior design. Helper agents run structured elicitation by generating targeted questions and drafting the tools/context/output artifacts in consistent Markdown for human validation, turning SME knowledge into structured engineering inputs.

**Phase 3: Reasoning Policy and Guardrails** This phase involves specifying the agent's interaction policy and verification behaviors, defining expected failure modes (e.g., uncertainty, missing evidence, tool errors, ambiguous queries), and establishing guardrails that determine when the agent should ask questions, refuse, escalate, or switch strategies. Helper agents draft candidate reasoning policies and guardrail behaviors and iteratively revise them based on SME and developer feedback, explicitly designing how the agent behaves. Phase 3 makes agent reasoning and safety behavior explicit and reviewable.

**Phase 4: Prompt Architecture and Tool-Orchestration Implementation** Phase 4 translates artifacts into an implementable agent prompt designed using patterns from the prompt catalog [2], grounding injection strategy, tool schemas, routing logic, and retry policies. The helper agent generates a prompt aligned with the input artifacts, and implementation is treated as a structured translation of artifacts into a working prompt for the agent. Consequently, Phase 4 treats the prompt as an engineered output derived from approved artifacts, not ad hoc iteration.

**Phase 5: Benchmarking and Verification** This final phase defines requirements for constructing a benchmark of realistic queries and defines scoring rubrics and pass/fail thresholds. In Phase 5, helper agents elicit responses from SMEs to identify potential sources that can be used to automate or semi-automate benchmark construction (e.g., types of queries, target answer types), and the resulting benchmark requirements artifact is then used by a separate workflow to generate the benchmark. For example, to create a data search benchmark, the auxiliary workflow can identify papers that can be used to draft different queries based on each paper and extract answers (i.e., data used) to those queries, after which evaluation can be integrated into the engineering loop.

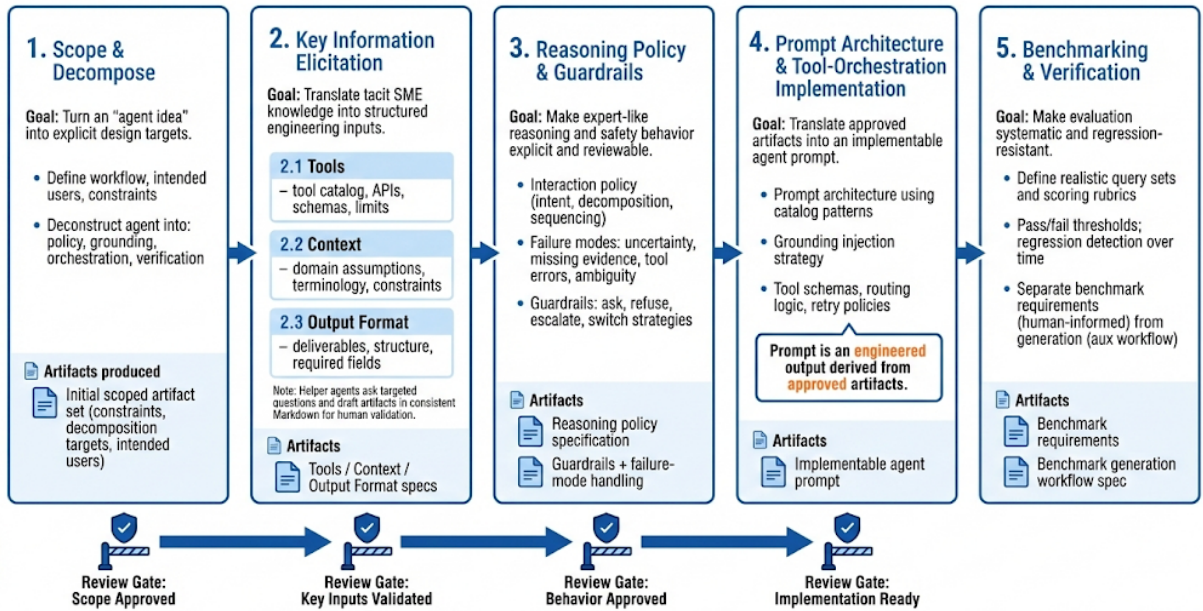

**Figure 2**: CARE methodology overview

It is important to note that helper agents use artifacts from previous phases to retain context, which allows the team to return to earlier phases and reiterate when an artifact does not pass a review gate without losing alignment or restarting elicitation from scratch. Artifact reuse across phases enables controlled iteration and continuity.

Helper agents make CARE repeatable at scale by continuously converting informal intent into structured Markdown artifacts and proposing candidate revisions while SMEs and developers retain authority through stage-gate approval. CARE's stage gates are defined by review and approval by both developers and SMEs of the phase artifacts, ensuring that the agent's intended behavior, constraints, and verification criteria remain aligned with real domain requirements and implementation feasibility.

CARE is intended to be model-agnostic and domain-agnostic because it focuses on producing portable design artifacts and verification practices that can be instantiated with different LLMs and applied across scientific and technical domains that require grounded, tool-using workflows. CARE artifacts and helper-agent prompt modules are treated as versioned engineering objects (e.g., stored and versioned in GitHub) so teams can audit changes, compare iterations, and relate specification and helper-agent prompt updates to changes in agent behavior observed during verification. Versioning connects evolving artifacts to measured outcomes and supports maintainability.

### 5.2 Helper Agents as Facilitation Infrastructure

CARE treats helper agents as required facilitation infrastructure that operates upstream of the deployed agent, because the central bottleneck in agent engineering is repeatedly translating

SME intent into precise, testable specifications that developers can implement. In CARE, helper agents facilitate structured elicitation by asking targeted questions aligned to the current phase's "information gathering" dimensions, capturing SME and developer answers, and converting them into concise Markdown artifacts that are legible to humans and usable as persistent context for downstream phases. Helper agents convert phase-specific elicitation into durable, reusable artifacts.

Minimum helper-agent capabilities include (i) accurately summarizing SME intent without introducing new requirements, (ii) asking structured, phase-aligned clarification questions when information is missing or inconsistent, (iii) drafting and editing concise Markdown artifacts using consistent templates and terminology, and (iv) proposing concrete revision candidates ("diffs") that SMEs and developers can accept, modify, or reject at review gates. CARE depends on helper agents that can reliably elicit, structure, and revise specifications. Helper agents also maintain continuity across phases by reusing artifacts from earlier stages as persistent context, which allows the team to revisit earlier phases when a review gate fails without losing alignment or re-running elicitation from scratch. Artifact continuity makes iteration controlled rather than chaotic.

Because helper agents drive elicitation, CARE's quality depends on whether helper agents ask questions that cover all relevant phase dimensions (e.g., tool constraints, provenance expectations, edge-case query types), and omissions can leave important failure modes unspecified and therefore untested. Elicitation coverage is a first-order risk, and CARE explicitly manages it via artifacts and gates. CARE constrains helper-agent influence through stage gates by requiring that SMEs and developers jointly review, correct, and approve the artifacts, ensuring that domain truth, safety constraints, and implementation feasibility remain human-owned even when helper agents drive drafting and iteration speed. Helper agents accelerate work, but humans control acceptance and accountability. The use of helper agents does introduce a distinct source of variability since artifact quality depends on how faithfully helper agents summarize, structure, and propose revisions. Therefore, CARE treats artifacts and helper-agent prompt modules as versioned objects and relies on verification gates to detect when updates change behavior in unintended ways.

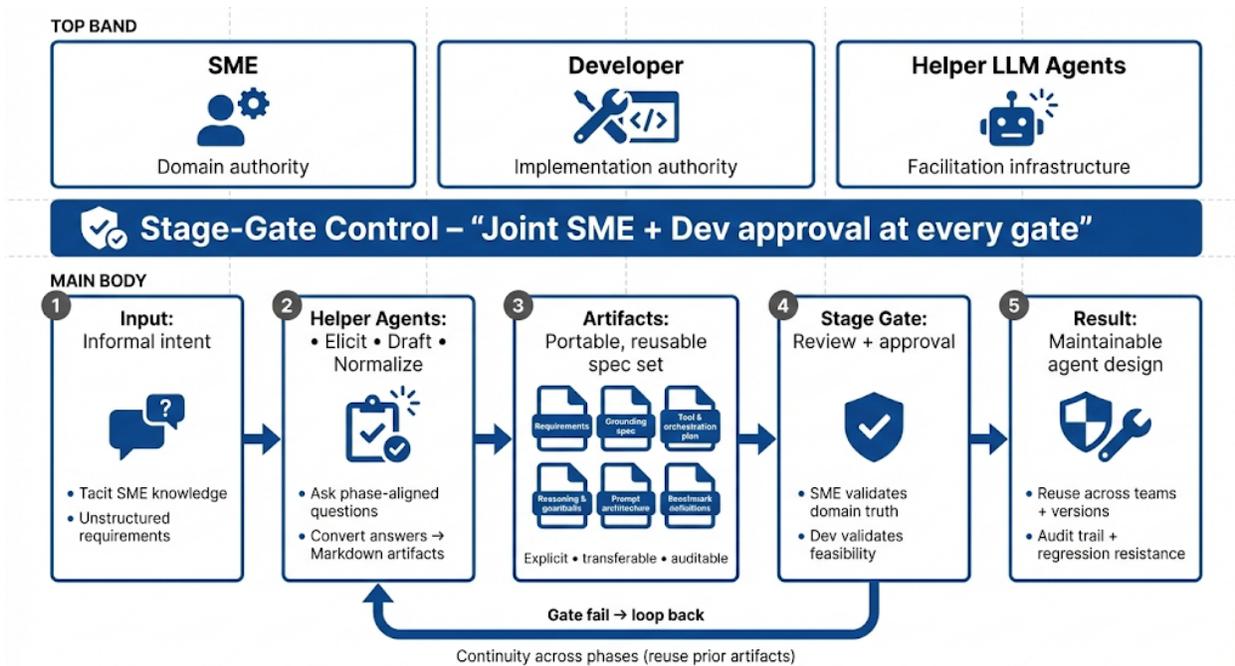

**Figure 3**: Three-party collaboration

## 6. Evaluation Case Study

### 6.1 Setup

We evaluate CARE through a case study in which we developed a NASA Earth science scientific data discovery agent, where the agent's purpose is to translate natural-language scientific data needs into actionable retrieval steps over an authoritative catalog. The agent retrieves datasets by invoking the NASA Common Metadata Repository (CMR) API, returning candidate dataset concept IDs as the retrieval outputs that downstream users can inspect and validate.

To isolate the impact of CARE's methodology rather than infrastructure differences, we compare two agents with identical model and tool access, differing primarily in whether their behavior and prompts were engineered through CARE's staged artifact process.

The first system, *cmr_care_v1*, is the CARE-designed agent, built from CARE artifacts (scope, tools/context/output, reasoning/guardrails, prompt architecture, benchmark requirements) and executed as a tool-using agent over the CMR API. The second system, *cmr_simple*, is a baseline agent that uses the same underlying model and the same CMR tool interface but does not incorporate CARE's staged artifact-driven design and gate process in the same way. The baseline represents a minimal "LLM + tool" approach for a fair comparison.

Both agents are evaluated under the same retrieval setting and output type (top-K retrieved concept IDs), so that measured differences reflect how the CARE-designed interaction policy, context handling, and tool-use behaviors influence retrieval performance. The setup controls the environment and focuses the comparison on behavior engineering.

### 6.2 Implementation

The case study implementation follows CARE as a staged sequence for designing a NASA Earth science data discovery agent. The process begins with the human workflow rather than the model or prompt so that design choices are anchored in real user constraints, decision points, and definitions of success. In each phase, helper agents drive the elicitation by asking targeted clarification questions organized around the phase's explicit "gather" dimensions, and they convert SME and developer answers into structured artifacts that are reviewed at the phase gate.

**Phase 1: Understand the user and the tasks (always first)** The first phase is executed with the science lead, developers, and SMEs, and helper agents ask questions to elicit who the user is (role, expertise, constraints), what tasks they are trying to accomplish, the workflow steps they typically follow, pain points, non-delegable decisions, desired outcomes, and definition of success.

**Phase 2: Key information elicitation** The next phase is performed before any reasoning or prompt design because the team cannot specify agent behavior without first knowing what actions the agent can take and what information it can reliably access. Helper agents structure questioning accordingly across tools, context, and outputs.

**Phase 2.1: Tools, data sources, and system constraints** These components are elicited through helper-agent questions that enumerate relevant tools, APIs, datasets, and resources; clarify input/output schemas, document limits, quotas, and permissions; and capture provenance and metadata requirements and constraints from policy, security, and governance.

**Phase 2.2: Context requirements** The requirements are elicited through helper-agent questions that define the agent's context access, retrieval strategies (metadata search, vector search, or hybrid), summarization rules, and memory boundaries. These inputs are then used to produce a context-window management plan that matches the intended reasoning strategy and eventual prompt implementation.

**Phase 2.3: Output formatting** Helper-agent questions help define structured output templates, citation and provenance expectations, rules governing when the agent should avoid giving final answers and instead ask the user to think, degradation behavior when data is incomplete, and supported output styles (e.g., narratives, tables, JSON).

**Phase 3: Reasoning policy and guardrails** These components are specified before prompt construction so that safety boundaries and the intended thinking logic determine what the prompt architecture must implement, and helper agents structure questions around guardrails and reasoning strategy rather than around prompt wording.

**Phase 3.1: Safety boundaries and guardrails** In this part of Phase 3, helper-agent questions capture forbidden actions, sensitive domains (e.g., embargoed data, human subjects, interpretation limits), what the agent must never guess or hallucinate, review/escalation requirements, and relevant ethical, organizational, and scientific norms, because guardrails must shape both the prompt architecture and the evaluation design.

**Phase 3.2: Reasoning strategy** Helper-agent questions define task decomposition logic, when the agent should ask questions, how it should retrieve, compare, critique, and synthesize, how it handles uncertainty, criteria for tool selection, and escalation rules such as abstain, ask the user, or flag an error, because the reasoning strategy is the blueprint that prompt patterns must implement.

**Phase 4: Prompt architecture and tool-orchestration implementation** This phase is executed only after the reasoning strategy and guardrails are defined. Helper agents generate a prompt architecture aligned to the approved artifacts using patterns from the prompt catalog, including persona specification, flipped-interaction prompts, planning prompts, critique/verification prompts, output patterns and formatting templates, tool-use scaffolding, and reflection/self-check instructions.

**Phase 5: Benchmarking and verification** This final phase defines what is needed to judge performance, and helper agents ask questions to elicit scenario-based tasks, test prompts, expected outputs, human-scored rubrics (scientific correctness, clarity, safety), a failure-mode catalog, and acceptance criteria. Benchmarks must reflect real tasks, real reasoning logic, and actual safety requirements.

### 6.3 Benchmarking and Metrics

The evaluation process involves a two-gate benchmarking approach to balance iteration speed and evaluation credibility, where a large synthetic benchmark provides rapid feedback and an SME-created gold benchmark provides higher-confidence validation. The two-gate design supports fast iteration without relying on synthetic data for the final claim.

In the first gate, we benchmark the CARE-designed agent against a baseline on a synthetic benchmark that can be produced quickly, and we treat this gate as a practical check that CARE does not underperform a minimal "LLM + tool" approach. The synthetic gate functions as an early warning signal for over-engineering. If the CARE-designed agent performs worse than the baseline on the synthetic gate, this suggests the CARE design may be misaligned or over-engineered and should be revisited. However, if the CARE-designed agent matches or exceeds the baseline, the evaluation proceeds to the gold benchmark where SME effort is higher but validity is stronger.

The second gate uses a gold benchmark created by SMEs as a smaller but higher-effort evaluation set of queries and expected answers, enabling final assessment of whether the CARE-designed interaction policy, constraints, and verification behaviors improve retrieval performance on realistic expert tasks. We evaluate retrieval quality using Recall@K, a standard information-retrieval metric that measures the proportion of expected (ground truth) datasets that appear within the top-K retrieved results returned by the agent. Recall@K provides a graded and widely accepted measure of retrieval success.

For a query with expected ground truth dataset concept IDs Expected and top-K retrieved dataset concept IDs Top-K Retrieved, Recall@K is computed as (|Expected ∩ Top-K,Retrieved| / |Expected|), yielding a score in ([0,1]) where higher is better and partial credit is awarded when an agent retrieves some but not all expected datasets. The metric captures "did we retrieve the right datasets within K," not only perfect ranking. We report Recall@1, Recall@3, and Recall@5 to characterize performance at different retrieval depths, reflecting practical usage scenarios where users may inspect only one or a few results. For each agent and benchmark condition, we compute mean Recall@K across queries and report the sample size (n) to contextualize the stability of estimates and the scale of each evaluation gate.

The synthetic benchmark is generated by ingesting Earth science research papers containing known dataset citations, using an automated LLM workflow to generate natural language queries grounded in paper context, and applying an automated validation loop that iteratively searches CMR until the expected dataset is retrieved. This process ensures queries are solvable and grounded in reality. Because each synthetic query is constructed to target a single specific dataset by design, Recall@1 is the primary metric for the synthetic gate, capturing whether the agent ranks the correct dataset as the top result. This evaluation design follows the principle that LLMs are useful for analysis when solving is hard but verifying is easy, and therefore emphasizes producing synthetic benchmarks that are easily human verifiable.

The gold benchmark consists of SME-formulated queries grounded in curated scientific publications that reference Earth science datasets, along with expected CMR concept IDs that serve as ground truth. The gold benchmark captures realistic expert intent and variability in query types.

### 6.4 Results

Performance of the CARE approach is measured using Recall@K (reported at K = 1, 3, 5), aggregated as mean Recall@K across queries for each benchmark condition.

The synthetic benchmark contains n = 621 queries and is constructed so that each query targets exactly one specific dataset by design, making Recall@1 the most important metric for this gate. The gold benchmark contains n = 43 SME-formulated queries grounded in curated scientific publications, with expected CMR concept IDs as ground truth and SME annotations such as difficulty and direct versus indirect query type. The gold benchmark tests performance on expert-authored, realistic queries. SMEs selected Recall@5 as a useful measure because as an end user they typically focus on the top five results. The CARE designed agent outperforms a baseline under identical model and tool access, achieving higher Recall@1 on the synthetic gate (n=621; 71.7% vs 69.1%) and higher Recall@5 on the gold gate (n=43; 27.2% vs 20.2%).

| Gate | Agent | Recall@1 | Recall@3 | Recall@5 |
|---|---|---|---|---|
| Synthetic (n=621) | cmr_care_v1 | 71.7% | 83.6% | 85.2% |
| | cmr_simple | 69.1% | 82.3% | 82.4% |
| Gold (n=43) | cmr_care_v1 | 7.8% | 22.6% | 27.2% |
| | cmr_simple | 9.7% | 15.6% | 20.2% |

**Table 1.** Two-gate evaluation results for NASA Earth science data discovery (Recall@K)

## 7. Discussions and Limitations

The case study suggests that CARE is most valuable when workflows require nuanced domain interpretation and constrained retrieval. CARE's staged artifacts excel at translating implicit SME practice into explicit reasoning policies, context/grounding requirements, tool-use constraints, and verification criteria that are otherwise difficult to keep consistent across iterations. As such, CARE should help most in domains where expert process discipline and constraint handling are essential. A key practical implication is that CARE's staged elicitation and artifact gates likely reduce coordination overhead by aligning SMEs and developers early on tools, context boundaries, output/provenance expectations, and guardrails, which reduces churn and rework during later prompt and orchestration implementation.

CARE's effectiveness depends on the quality of helper agents, particularly their ability to ask the right questions across the relevant "information gathering" dimensions in each phase and to convert answers into faithful artifacts without introducing new requirements or silently omitting constraints. CARE also depends on the expertise levels of SMEs and developers, because SMEs must surface nuanced domain constraints and validate artifact fidelity while developers must ensure tool realism and feasibility. Weaknesses in either role can propagate into artifacts, prompts, and evaluation design even when the process is followed.

CARE mitigates helper-agent and human variability risks through stage gates in which SMEs and developers jointly review, correct, and approve artifacts, ensuring that domain truth, safety constraints, and implementation feasibility remain human-owned even when helper agents accelerate drafting. The reported performance improvements should be interpreted cautiously because benchmark choice, query distributions, and metric definitions can materially affect conclusions, and alternative benchmark constructions or different choices of K could change the observed magnitude of gains. A potential drawback to validity is benchmark leakage or overfitting that can occur through multiple pathways, including iterative refinement of artifacts and prompts to the benchmark distribution, helper-agent elicitation that implicitly "teaches to the test". This risk is compounded if helper-agent elicitation fails to cover important dimensions (e.g., provenance expectations, edge-case query types, tool constraints) or if stage-gate reviews become superficial, because teams may optimize benchmark-visible behaviors while leaving critical real-world failure modes unspecified and therefore untested.

Model and provider drift is an operational limitation because changes in model behavior can alter compliance with reasoning policies and guardrails and can also affect helper-agent drafting behavior, motivating periodic re-verification through CARE's stage gates and controlled versioning of artifacts and helper-agent prompt modules.

Although the case study is tool-using, CARE is applicable to both tool-using and no-tool agents because its core value is specifying interaction policy, grounding, and verification artifacts, while orchestration artifacts can be minimal or omitted when external tools are not part of the workflow.

## 8. Conclusion

This paper introduced CARE (Collaborative Agent Reasoning Engineering), a disciplined, stage-gated methodology for engineering LLM agents that translates domain expertise and workflow requirements into reusable artifacts specifying interaction policy, grounding, tool orchestration, and verification rather than relying on trial-and-error prompt tinkering. CARE is positioned as a specification-first methodology for building maintainable agents. Implementation of CARE involves a three-party collaboration among SMEs, developers, and required helper agents, where helper agents facilitate structured elicitation, draft and normalize artifacts, and maintain continuity across phases while SMEs and developers retain authority through artifact review and approval at stage gates. Dependable agent behavior requires explicit specifications and interfaces, because failures often arise at the boundaries between design targets and remain difficult to detect under ad hoc prompting.

In a NASA Earth science data discovery case study using the NASA CMR API, we evaluated a CARE-designed agent against a baseline under identical model and tool access using a two-gate approach that first benchmarks on a large synthetic dataset for rapid iteration and then validates on an SME-created gold benchmark for higher-confidence assessment. Using Recall@K as the retrieval-quality metric, the CARE-designed agent outperformed the baseline on both gates, including higher Recall@1 on the synthetic benchmark and higher Recall@5 on the SME-created gold benchmark. CARE shows consistent retrieval gains under both fast synthetic testing and SME-grounded verification.

Beyond measured retrieval quality, CARE provides a repeatable engineering workflow by enforcing a staged progression from scope and elicitation to reasoning/guardrails to prompt architecture and finally benchmarking/verification, while treating artifacts and helper-agent prompt modules as versioned engineering objects (e.g., stored and versioned in GitHub) to support auditability and maintainability. CARE's process discipline and versioned artifacts support controlled iteration and long-term maintenance.

Implementation of CARE faces several key limitations, including benchmark sensitivity, risk of multi-pathway leakage, and heavy dependence on the quality of helper-agent elicitation. Furthermore, variability in expert knowledge and model drift can degrade performance affecting both the deployed agent and the helper-agent facilitation layer, which is why periodic re-verification through CARE's stage gates is essential for maintaining accuracy.

Future work will prioritize multi-domain replication and holdout-style transfer evaluation to test how CARE performs across domains, query distributions, and tool ecosystems, and to identify which phases and artifacts most strongly drive improvements in retrieval quality, safety behavior, and maintenance cost.